\documentclass[runningheads]{llncs}
\usepackage{graphicx}
\usepackage{subfig}
\usepackage{fancyvrb}
\usepackage{hyperref}
\usepackage{comment}
\usepackage{cite}
\usepackage{tabularx}
\usepackage{tikz} 
\usetikzlibrary{shapes,backgrounds,calc,arrows, graphs}
\tikzset{main node/.style={circle,fill=blue!95,draw,inner sep=2.5pt,text=white},
        comm node/.style={circle,fill=red!95,draw,inner sep=2.5pt,text=white},
        summ node/.style={rectangle,fill=red!75,draw,inner sep=5pt,text=white},
            }
\newcommand\blfootnote[1]{%
  \begingroup
  \renewcommand\thefootnote{}\footnote{#1}%
  \addtocounter{footnote}{-1}%
  \endgroup
}
            
\begin{document}
\title{Distinguishing\\
Commercial from Editorial Content in News}
%
%
\author{Timo Kats\inst{1}\orcidID{0000-0003-1650-1814} \and
Peter van der Putten\inst{1}\orcidID{0000-0002-6507-6896} \and
Jasper Schelling\inst{2}\orcidID{0000-0002-9995-1505}}
%
%
\institute{Leiden University, Niels Bohrweg 1, 2333 CA Leiden, The Netherlands\\ \email{t.p.a.kats@liacs.leidenuniv.nl, p.w.h.van.der.putten@liacs.leidenuniv.nl} \and
Stichting ACED, Amsterdam, The Netherlands\\ \email{jasper@aced.site}}
\maketitle              

\begin{abstract}
How can we distinguish commercial from editorial content in news, or more specifically, differentiate between advertorials and regular news articles? An advertorial is a commercial message written and formatted as an article, making it harder for readers to recognize these as advertising, despite the use of disclaimers. In our research we aim to differentiate the two using a machine learning model, and a lexicon derived from it. This was accomplished by scraping 1.000 articles and 1.000 advertorials from four different Dutch news sources and classifying these based on textual features. With this setup our most successful machine learning model had an accuracy of just over $90\%$. To generate additional insights into differences between news and advertorial language, we also analyzed model coefficients and explored the corpus through co-occurrence networks and t-SNE graphs.

\end{abstract}

\section{Introduction}
\blfootnote{Paper presented at 33rd Benelux Conference on Artificial Intelligence and the 30th Belgian Dutch Conference on Machine Learning (BNAIC/BENELEARN 2021), Luxembourg, November 10-12, 2021}

In journalism it is best practice to clearly distinguish between editorial and sponsored commercial content. This is referred to as the `separation of church and state' in media \cite{Connill}. However, some forms of advertising have made this separation less clear to readers and therefore threaten this principle.

An example of this is the advertorial, which is commercial content in the form of an article. Advertorials are an example of what marketers call `native advertising'. In fact, advertorials are so much like articles, that despite using disclaimers and different layouts most readers don't notice the difference. In a study conducted by the university of Georgia only 8\% of readers recognized advertorials as commercial content \cite{Wojdynski}. As a result of this, advertorials have made the separation of church and state in the news less clear. 

That's why this research aims to differentiate articles and advertorials using machine learning. We would like to answer two research questions. Firstly, to what extent can we differentiate commercial and editorial content by a using machine learning model, and a lexicon derived from this? Secondly, can we use AI and machine learning to better understand the difference between commercial and editorial language? 

It's important to note that the separation of commercial and editorial content is hotly debated in journalism and the society at large. Yet, to our knowledge a machine learning  based perspective to identify advertorials and commercial messaging was not part of this debate yet. By doing this we not only hope to answer our research questions, but also showcase how machine learning can be a solution in the debate surrounding the usage of advertorials. 
This research has been carried out in the context of the Reverb Channel program \cite{Schelling2020}, a data driven exploration of our networked news culture that aims to reverse the sometimes questionable role of AI in digital media, by using it to investigate topics such as framing, polarization and ideology spaces.\footnote{\url{https://www.aced.site/en/programmes/reverb-channel}}

The remainder of this paper is structured as follows. Section 2 provides more background and related work. Section 3 explains the process of acquiring our data, followed by sections on our classification approach, and on our exploratory co-occurrence network based approach to increase the insight into how language differs across advertorials and news. Section 6 concludes the paper.

\section{Background and Related Work}

Even though we are not aware of any other research to leverage machine learning to distinguish advertorials from editorial content, the discussion around the usage of advertorials and commercial content in general is broader than this research alone, and has been debated widely in journalism and marketing. In this section we discuss some of this background context.

\subsection{The change of journalism’s business model in the digital age.}

The rise of the internet has had a lot of effect on journalism. It opened up a whole new channel for news content, but it also it negatively impacted circulation and advertising revenue for traditional news channels. 
For example, US weekend circulation of newspapers declined from 59.4 million (2000) to est. 25.8 million (2020), revenue from advertising declined from 48.7 billion (2000) to est. 8.8 billion (2020), whilst revenue from circulation remained relatively stable (10.5 (2000) to 11.1 (2020)), and the share of advertising revenue increased from $17\%$ (2011) to $39\%$ (2020) \cite{Pew2021}. So despite drastic drops in circulation, companies were able to protect circulation income, but advertising revenues dropped dramatically. These developments altered the business model of journalism significantly, and drove publishers to find new sources of advertising revenue, such increased usage of advertorials and other forms of sponsored content.

\subsection{Disguise, deception and disclosure in advertorials.}

As discussed, whilst in journalism the distinction between editorial and sponsored commercial content is a key principle, this is challenged by advertorials in practice as readers have a hard time differentiating these from editorial content, despite the use of labels and disclaimers \cite{Connill}. 

Advertorials can be both deceptive and effective. As a classical example, in 1989 the R.J Reynolds Tobacco Company had settled charges with the FTC on that it had made false and misleading claims in an advertorial on health effects of smoking, titled `Of cigarettes and science'. Wilkinson et al. subsequently ran a test and over a quarter of participants thought the article was editorial content, not commercial \cite{Wilkinsonsetal1995}. 

In another study by Kim et al., the use of an advertorial over a standard advertisement increased the relevance of and attention to the message, and message and elaboration and recall. It made no difference whether the advertorials were labeled as such, and over two thirds of subjects exposed to labeled advertorials were not able to recall whether these advertorials were labeled or not \cite{Kim2001}. 

As mentioned in the introduction, in another study by the University of Georgia only 8\% of readers recognized advertorials as commercial content \cite{Wojdynski}. In their study, the use of disclaimers did have a positive impact on recognizing the text as commercial, with best effects for placement of disclaimers in the middle or the bottom, and explicit use of words such as `advertizing' and 'sponsored'. Also Krouwer et al. found that small changes, such as the location of a disclaimer, significantly impacts the recognizability for readers \cite{Krouwer}. Apart from readers not noticing labeling, advertorials often violate guidelines for labeling, formatting and content \cite{Cameron1996}.

To provide perhaps a somewhat more positive view on advertorials, in a survey by Reijmersdal et al. of subscribers of Dutch magazines, when asked explicitly only 12\% of respondents thought advertorials are deceptive \cite{Reijmersdal2005}. 

 The more established newspapers and magazines will make more of a serious effort to make it known that certain content is sponsored, and writers producing advertorials are kept separate from the editorial teams. But is that sufficient, also when taking the proliferation of new digital media titles and the ongoing pressure to increase advertizing revenues into account, and norms are shifting towards further integration between editorial and commercial teams and objectives \cite{Cornia2020}? 
 
 The results above may vary but in our opinion this is clearly not sufficient. The ability to disguise content, willingly or unwillingly, and the probability that advertorials are not recognized as such even if properly labelled is significant. Marketers call it native advertizing for a reason.
 
 The risk of mistaking commercial content for objective editorial content is somewhat obvious, but note there can be an opposite detrimental effect as well. For instance, Iversen at al. observed that exposure to native political ads reduced the public's trust in political news \cite{Iversen2019}.
 
\subsection{The usage of lexicons in classifying text}

As mentioned, we aim to create a classification model and lexicon that distinguishes editorial from commercial language. Whilst text classification models are used abundantly in NLP research, we are also looking to distribute our artifacts to journalists and other non-technical audiences. In domains such as social science lexicons are often used, for common tasks such as sentiment analysis \cite{Taboada} or more specific tasks, such as detecting moral foundations in ethical reasoning \cite{graham2009liberals, Teernstra2016}. Lexicons can be handcrafted or created through linguistic analysis, and typically include keywords that indicate a particular class, potentially including a weight.

We were not able to identify prior work that uses machine learning, handcrafted or trained lexicons to differentiate advertorials from editorial content. A different, yet relevant related work is the study by Zhou, who uses genre analysis to characterize the general structure and linguistic characteristics of advertorials, using mostly manual analysis and interpretation \cite{Zhou2012}.

\section{Data Acquisition}

In order to make a model that answers the research questions mentioned earlier we have created a data set with advertorials and regular news articles. The Reverb Channel corpus contains millions of articles \cite{Schelling2020}, but no advertorials, hence we had to acquire our own data for this research. In this section we explain this process and showcase the data set that we acquired. For full details we refer to \cite{Kats2021}.

\subsection{Scraping the data}

The data for this research had to be scraped directly from news sources using web crawlers. For our research we used Python and the BeautifulSoup library. With this set up we made a URL-scraper and a web-scraper for every news source. We first collected the URLs from the pages we wanted to scrape data from and thereafter use those URLs to collect all the data we needed with the web-scraper. We also carried out additional cleaning and transformation, such as removal of all commas, translation of any HTML to flat text where needed, and lowercasing of all text.\footnote{Source code, the lexicon and other deliverables can be found at 
\url{https://github.com/TimoKats/research_distinguishing_commercial_and_editorial_content}}

\subsection{Resulting data set}

The data set that we acquired with this method has 2000 entries in total, about half of these entries are advertorials (see Figure 1). These entries are roughly equally distributed over four different news sources. These news sources are (online-only news) Nu.nl, (politically conservative) Telegraaf, (politically progressive) NRC, and (business publication) De Ondernemer. By including these four different news sources with roughly equal number of documents in the data set we strive to create an unbiased data set that is representative of the Dutch media landscape as a whole. 

\begin{figure}[tp]
    \centering
    \subfloat{\includegraphics[width=5.5cm]{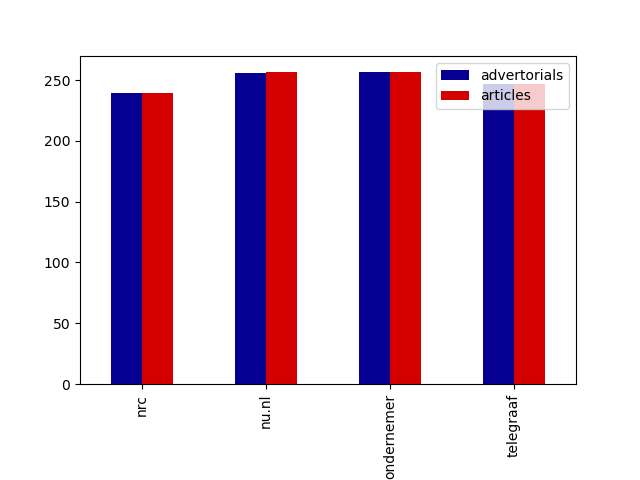}}
    \qquad
    \subfloat{\includegraphics[width=5.5cm]{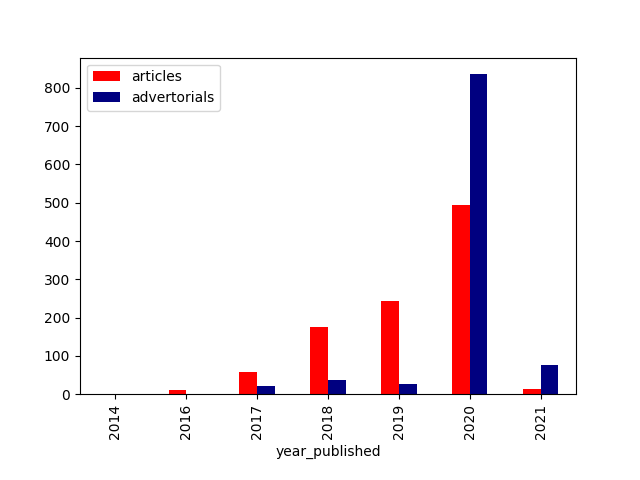}}
    \caption{Metadata from the acquired data set}
\end{figure}

\section{Distinguishing Advertorials from Regular Articles with Classification Models}

With this corpus we developed classification models and a corresponding lexicon, which also gave us some first insights into differences between the language used in advertorials versus news.

\subsection{Experimental set up}

In terms of cleaning the data, we first removed potential leakers. Leaking variables in our model refer to words that trigger the model whilst being unique to our data set and media covered, for example sponsor names and disclaimers. To further lower the risk of leakage, we excluded the title and focused on the main text. Furthermore, we experimented with regular bag of words (BoW) as well as TFIDF weighted BoW, the removal of stop words and the number of features. Obviously, we could have easily obtained classifiers with near perfect performance, for instance by including disclaimer texts, but we were primarily interested in models that could distinguish commercial from editorial language.

For modeling, we selected a diverse set of classification methods to experiment with: SVM (default with rbf kernel), linearSVC, decision tree, random forest, k-NN, SGD and naive bayes. We restricted ourselves to these more classical methods as opposed to deep learning methods such as BERT, given that our data sets are relatively small, and interpretability of the results is key, for instance to iteratively identify leakers and get more insights into the difference between text types. 

To simplify the approach, we aim to find the best performing model (incl. parameter optimization) through narrowing down the search as the experiment progresses, taking the best performing preliminary results and continuing to optimize it. A limitation of such an approach is that the estimate of final accuracy may be somewhat optimistic given the sequential nature of the experiments (manual overfitting), but a full multidimensional experimental set up was too computationally expensive, and the scarcity of advertorials limited the use of an additional hold out test set. This could be addressed in future work. 

For SVM, SGD and linearSVC we increased the maximum amount of iterations to 5000 and for decision tree and random forest we set the max depth to ``none". In terms of evaluation we ran 10-fold cross validation to test various algorithms and parameters, as well as a cross domain test set up where one medium is used as the test set, and models are trained on the other media. The metrics that we evaluate our results with are accuracy, f1 score and AUC. 

\subsection{Results}

In a first set of experiments we benchmark the performance of all algorithms across regular and TF-IDF weighted BoW representations. Table~\ref{table_allalgsnostopwords} shows the results, with stop words removed; the results with stop words included were very similar. TF-IDF typically outperformed regular BoW so the remainder of the experiments was carried out with TF-IDF, with stop word filtering.
\begin{table}[tp]
\centering
\begin{tabular}{|l|l|l|l|l|}
\hline
\textbf{representation} & \textbf{learning model} & \textbf{accuracy} & \textbf{f1 score} & \textbf{auc} \\ \hline
bag of words            & svm                     & 0.85±0.04         & 0.85±0.05         & 0.93±0.04         \\ \hline
bag of words            & linearSVC               & 0.84±0.05         & 0.84±0.06         & 0.9±0.05          \\ \hline
bag of words            & decisionTree            & 0.78±0.08         & 0.78±0.08         & 0.79±0.07         \\ \hline
bag of words            & randomForest            & 0.88±0.06         & 0.89±0.07         & 0.94±0.05         \\ \hline
bag of words            & k-NN                    & 0.57±0.14         & 0.63±0.12         & 0.58±0.16         \\ \hline
bag of words            & SGD                     & 0.87±0.07         & 0.86±0.08         & 0.93±0.05         \\ \hline
bag of words            & naiveBayes              & 0.76±0.11         & 0.77±0.09         & 0.76±0.11         \\ \hline
tfidf                   & svm                     & 0.89±0.05         & 0.89±0.05         & 0.94±0.04         \\ \hline
tfidf                   & linearSVC               & 0.91±0.05         & 0.91±0.05         & 0.95±0.03         \\ \hline
tfidf                   & decisionTree            & 0.78±0.07         & 0.79±0.07         & 0.8±0.07          \\ \hline
tfidf                   & randomForest            & 0.88±0.07         & 0.89±0.06         & 0.94±0.05         \\ \hline
tfidf                   & k-NN                    & 0.51±0.03         & 0.64±0.02         & 0.51±0.05         \\ \hline
tfidf                   & SGD                     & 0.9±0.05          & 0.9±0.06          & 0.95±0.04         \\ \hline
tfidf                   & naiveBayes              & 0.76±0.09         & 0.76±0.08         & 0.76±0.1          \\ \hline
\end{tabular}%
\caption{Cross-validation accuracy with removal of stop words}
\label{table_allalgsnostopwords}
\end{table}

The results of the cross domain testing experiment can be found in Table~\ref{table_crossdomain}. The best results were obtained with SVM, linearSVC, random forest and SGD, closely followed by decision trees and naive bayes, and k-NN scored poorly, probably due to high dimensionality. Top scoring results were close, but SVM scored best, so we decided to continue the experiments with this method. In terms of media, NRC scored best, followed by Nu.nl and Telegraaf, and Ondernemer scoring substantially worse, which may be due to the fact that in the business to business domains editorial and commercial content is more similar.
\begin{table}[tp]
\centering
\begin{tabular}{| l | l | l | l | l | l | l | l | l | l |}
\hline
\footnotesize
& SVM   & Linear    & Decision  & Random & k-NN & SGD   & Naive &   \\
&       & SVC       & Tree      & Forest &      &       & Bayes & \\ \hline
Nu.nl       & 0.84& 0.84      & 0.72      & 0.83  & 0.52  & 0.84  & 0.72  & $0.76 \pm 0.12$ \\ \hline
NRC         & 0.93 & 0.95 & 0.75 & 0.82 & 0.52 & 0.95 & 0.81  & $0.82 \pm 0.15$\\ \hline
Ondernemer  & 0.76 & 0.68 & 0.54 & 0.55 & 0.51 & 0.65 & 0.56  & $0.61 \pm 0.09$\\ \hline
Telegraaf   & 0.85 & 0.84 & 0.66 & 0.84 & 0.46 & 0.83 & 0.73  & $0.74 \pm 0.14$\\ \hline
& 0.85 & 0.83 & 0.67 & 0.76 & 0.5 & 0.82 & 0.71 &  \\ 
& $\pm 0.06$ & $\pm 0.10$ & $\pm 0.08$ & $\pm 0.12$ & $\pm 0.02$ & $\pm 0.11$ & $\pm 0.09$ & \\ \hline 
\end{tabular} 
\caption{Cross-domain testing results (test set in rows, trained on other media, metric is accuracy)}
\label{table_crossdomain}
\end{table}

We also ran a structured experiment where we gradually increased the number of features that made clear that at 5000 features performance more or less stabilizes (results omitted for brevity), and we ran a series of tests to study the impact of tweaking the various SVM parameters (Table~\ref{table_svm}). 

\begin{table}[p]
\centering
\begin{tabular}{|l|l|l|l|l|}
\hline
\textbf{kernel} & \textbf{decision function} & \textbf{accuracy} & \textbf{f1\_score} & \textbf{roc\_auc} \\ \hline
linear          & ovo                                & 0.9029±0.0559     & 0.9003±0.0581      & 0.9495±0.0341     \\ \hline
linear          & ovr                                & 0.9029±0.0559     & 0.9003±0.0581      & 0.9495±0.0341     \\ \hline
poly            & ovo                                & 0.8094±0.0774     & 0.7755±0.1016      & 0.9167±0.0412     \\ \hline
poly            & ovr                                & 0.8094±0.0774     & 0.7755±0.1016      & 0.9167±0.0412     \\ \hline
rbf             & ovo                                & 0.8999±0.0579     & 0.8989±0.0583      & 0.9438±0.0393     \\ \hline
rbf             & ovr                                & 0.8999±0.0579     & 0.8989±0.0583      & 0.9438±0.0393     \\ \hline
sigmoid         & ovo                                & 0.9009±0.0563     & 0.8986±0.0585      & 0.9498±0.0339     \\ \hline
sigmoid         & ovr                                & 0.9009±0.0563     & 0.8986±0.0585      & 0.9498±0.0339     \\ \hline
\end{tabular}
\caption{The effect of tweaking the parameters with svm}
\label{table_svm}
\end{table}

\begin{figure}[tp]
    \centering
    \subfloat{{\includegraphics[width=5.5cm]{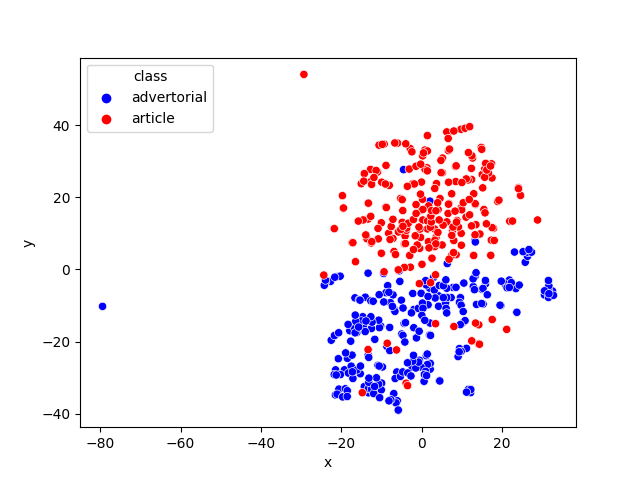}}}
    \subfloat{{\includegraphics[width=5.5cm]{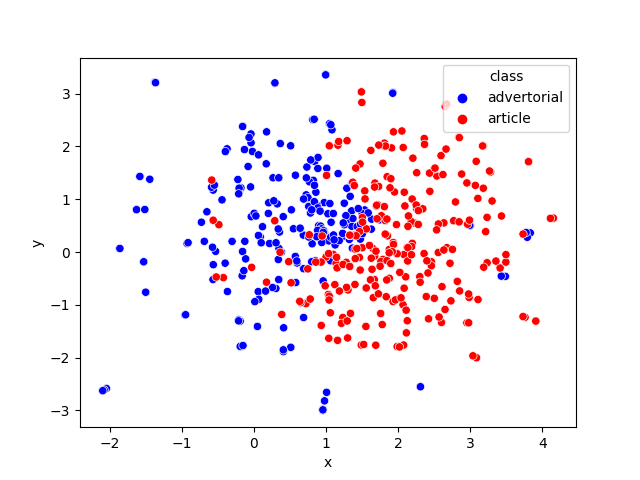}}}
    \qquad
    \subfloat{{\includegraphics[width=5.5cm]{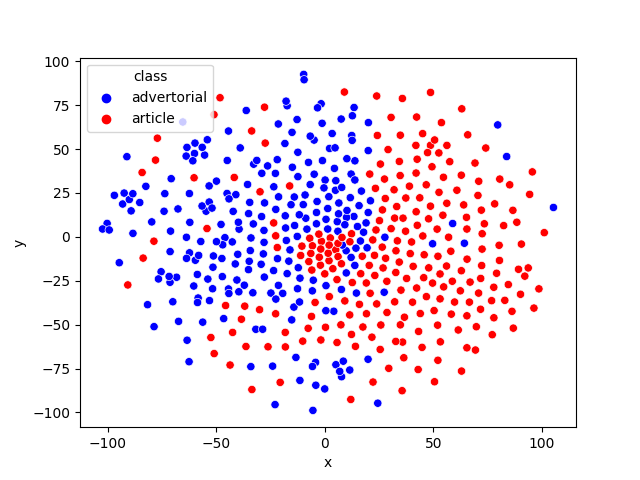}}}
    \subfloat{{\includegraphics[width=5.5cm]{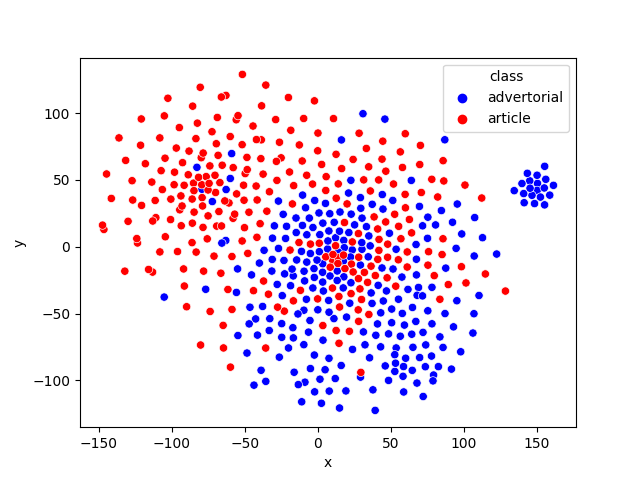}}}
    \caption{t-SNE plots for NRC (95\%), Nu.nl (92\%), Telegraaf (91\%) and Ondernemer (85\%)}
    \label{figure_tsne}
\end{figure}

To further validate the cross domain results, we also trained and tested models with data from just one medium each, and created t-SNE graphs (Figure~\ref{figure_tsne}, along with the corresponding accuracies). t-SNE graphs \cite{Maaten} are a way to represent multi-dimensional data (in our case a 5000 dimensions) in a two-dimensional scatter plot. For our experiment, we ran the t-SNE graph with a perplexity of 30, a maximum number of iterations of 1000 and a random state of 2. In other words, apart from the random state only the default parameter values. \\
Using this method we can visualize how well the classes can be separated based on the available data, making it possible to visualize the separation of church and state in our experiment. The ranking of various media are consistent with the cross domain results, with NRC displaying the clearest separation and Ondernemer the worst. 

\begin{table}[tp]
\centering
\resizebox{\textwidth}{!}{%
\begin{tabular}{|l|l|l|l|l|l|}
\hline
\textbf{learning model} & \textbf{features} & \textbf{text representation}  & \textbf{kernel} & \textbf{max iter} & \textbf{accuracy}  \\ \hline
svm                     & 5000              & tf-idf                        & linear          &  5000                             & 0.9029±0.0559      \\ \hline
\end{tabular}%
}
\caption{Settings from the final model.}
\label{table_finalmodel}
\end{table}

After completing the experimental process explained earlier we found that the model explained in Table~\ref{table_finalmodel} gave us the best results (all other parameters are defaults). So we used this model to derive a lexicon by training a model on all data and using this model's feature terms and weights. 
This is useful, even though it serves the same purpose as our model, because it can be published without publishing the data as well, which we are not able to do because of copyright issues, and it can be consumed more easily by a broad non technical audience such as journalists and social scientists. \footnote{The lexicon is published at \url{https://github.com/TimoKats/research_distinguishing_commercial_and_editorial_content}}

\begin{figure}[tp]
    \centering
    \includegraphics[width=0.8\textwidth]{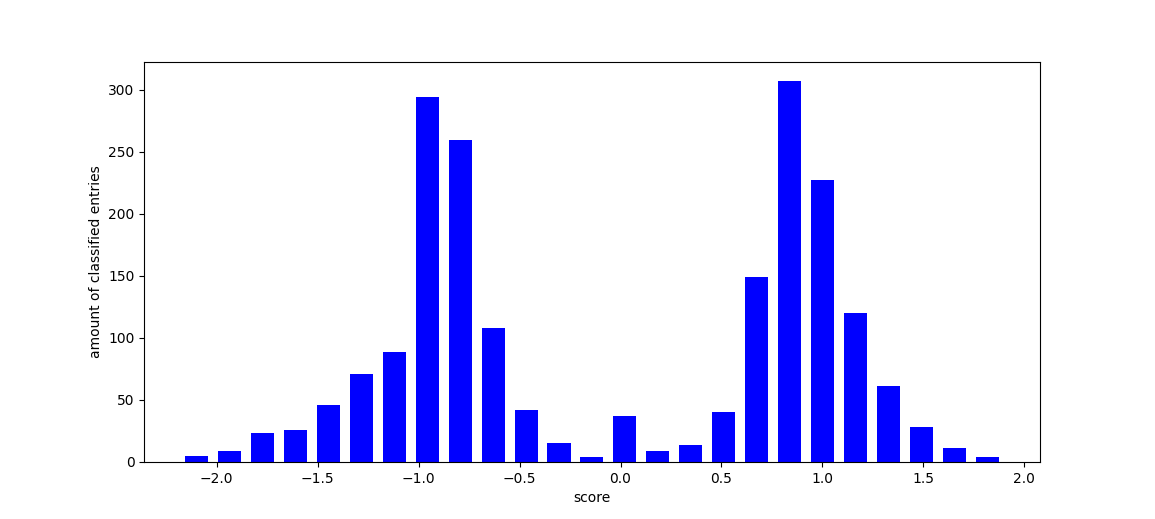}
    \caption{The distribution of the predicted scores from the entries using the lexicon.}
    \label{figure_lexicon}
\end{figure}

Using a linear kernel means that the separating hyperplane is defined in the original input space, hence we can interpret the weights of the model as term weights in a lexicon. Users can make very simple use of the lexicon, just by counting the occurrence of negative and positive words (with zero as threshold) or approximate the original model closer, for example by calculating a score by multiplying frequency of the terms with the term weights and summing the results. Figure~\ref{figure_lexicon} shows the distribution of the scores  for the full corpus, calculated with the latter approach. One can clearly see two more or less normal distributions representing the advertorials and regular articles.

\begin{figure} [tp]
    \centering
    \includegraphics[width=\textwidth]{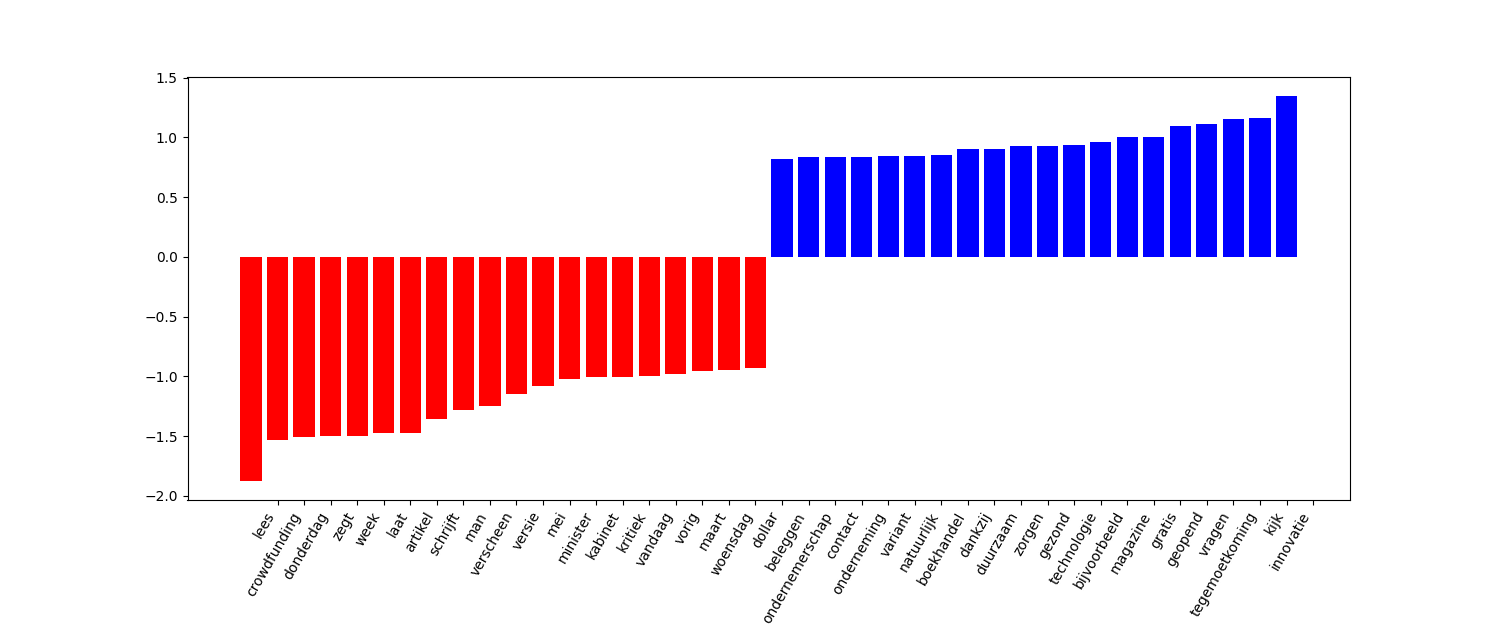}
    \caption{Coefficients with highest absolute values for regular articles (left) and advertorials (right)}
    \label{figure_featureimportance}
\end{figure}

Inspection of these feature coefficients also provides further insight into differences in language use between classes. In Figure~\ref{figure_featureimportance} we have listed the features with the highest absolute values for regular articles and advertorials. As can be seen, the difference isn't just a matter of topics (f.i. `cabinet', `minister' for news, `investing', `enterprise', `technology', `innovation' for advertorials), but also a matter of how these topics are being talked about. In regular articles, indications of time (days, months etc) and attribution (`writes', `says', `appeared') score high, whereas high scoring features for advertorials include adjectives such as `free', `healthy' and `sustainable', perhaps highlighting the benefits of products and services. Coefficients for the full 5000 features can be downloaded with the lexicon, and we also investigated by training models on media separately to understand differences between publications.

\section{Exploring the Corpus with Co-occurrence Networks}

\begin{figure}[tp]
    \centering
    \includegraphics[width=0.8\textwidth]{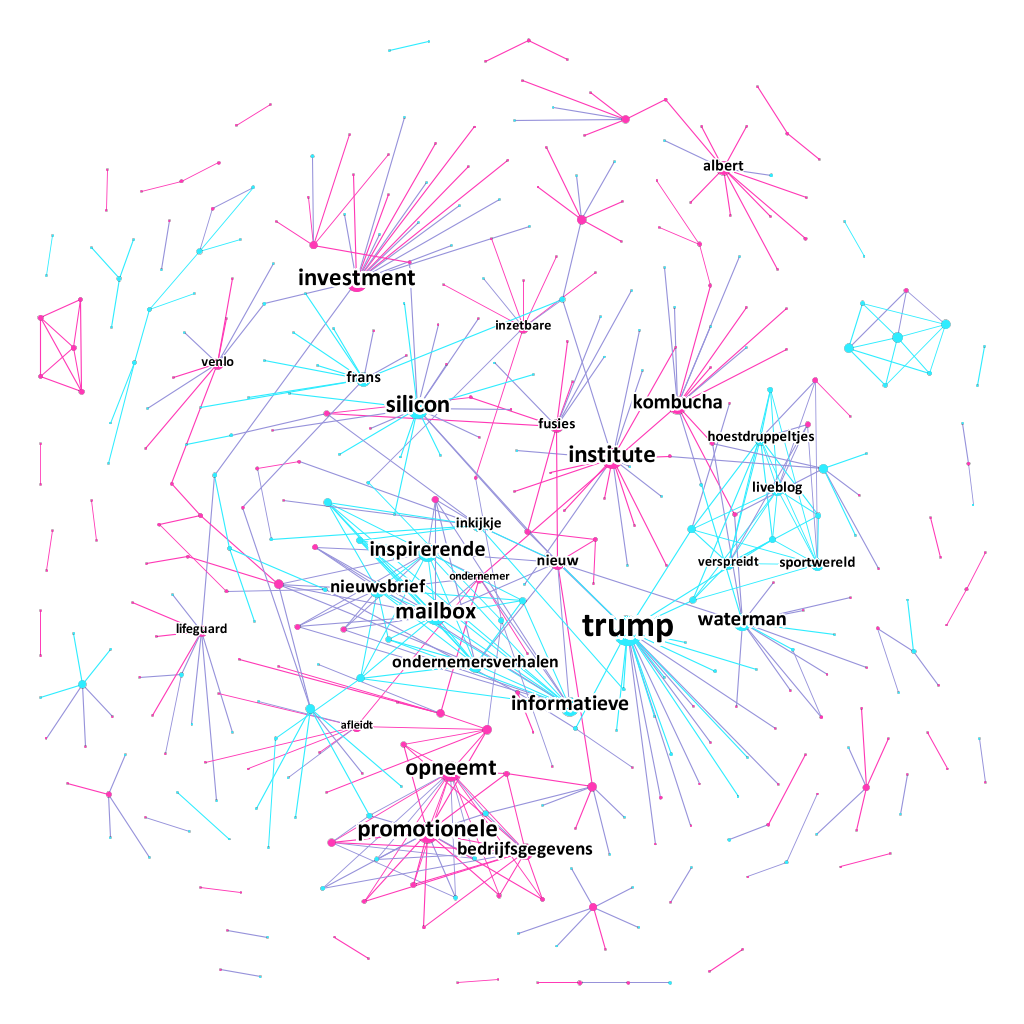}
    \caption{Overview of the co-occurence network. Complete version can be found at: \url{timokats.github.io/network/}}
    \label{figure_graph}
\end{figure}

Results such as feature importance already provided us with some insights into how language differs between advertorials and regular articles, but to delve deeper in a more exploratory fashion we have also created co-occurrence networks (see Figure~\ref{figure_graph}). 

\begin{figure}[tb]
    \centering
    \subfloat{{\includegraphics[width=0.45\textwidth]{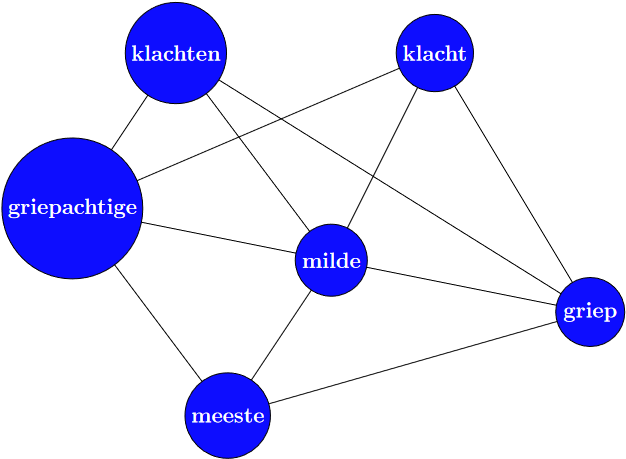}}}
    \hspace{0.08\textwidth}
    \subfloat{{\includegraphics[width=0.45\textwidth]{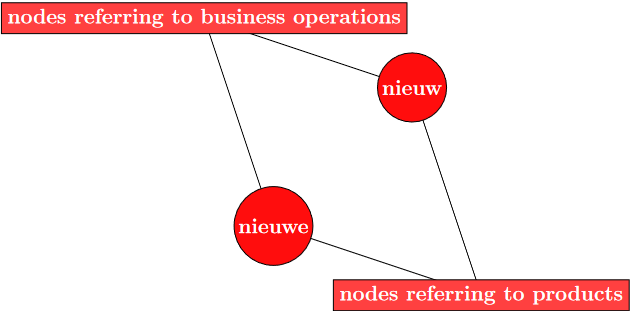}}}
    \qquad
    \qquad
    \subfloat{{\includegraphics[width=0.47\textwidth]{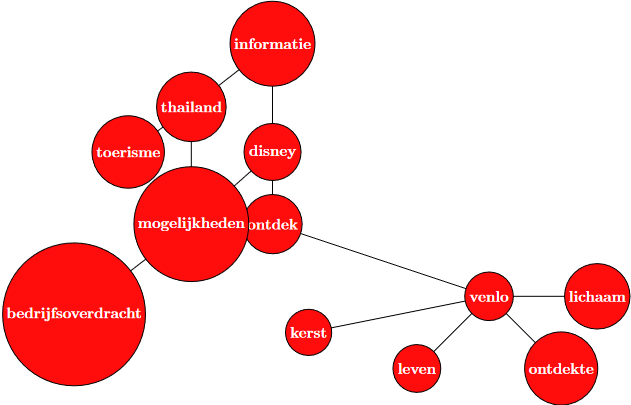}}}
    \hspace{0.08\textwidth}
    \subfloat{{\includegraphics[width=0.44\textwidth]{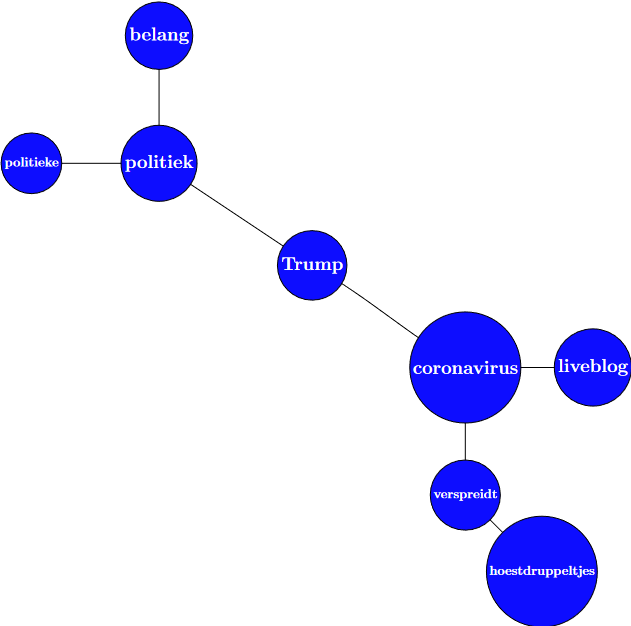}}}
    \caption{Commercial and editorial sub graphs.}
    \label{figure_subgraphs}
\end{figure}

The nodes in this network are the terms from the lexicon, blue nodes are the editorial terms and the red nodes are the commercial terms (negative and positive coefficients). The size of the node is related to its degree. The edges represent the connection between two terms in the data set. We calculated this based on how often the two terms appear in the same sentence as a percentage. So for example, in our data set every time the term `artificial' appears in a sentence, 75\% of the time that sentence also has the word `intelligence'. Thus, there's a directed edge from `artificial' to `intelligence'. For visualization we show all nodes with edges exceeding a minimum threshold, or likewise this could be seen as an undirected graph where the weight is the lowest value for the two terms.

By exploring the co-occurrence network certain things about both our results and data become apparent. First, the fact that some of our data centers around subjects that were very prevalent in 2020 (like the US elections and the covid-19 pandemic), resulting in a time frame bias towards 2020, because in future implementations of our model and/or lexicon these subjects may be less prevalent. 

Second, it has also given us more insight into the structure of commercial language and how it's different from editorial language. For example, commercial language in our network has two large clusters (one related to goods and one related services). These clusters are linked by the terms `nieuw' (new) and `nieuwe' (new). For our editorial clusters we for example found a cluster related to covid symptoms, which showcases the time frame bias mentioned earlier. Through using a co-occurrence graph we can find patterns/clusters like these and gain more insight into our data and results. An overview of some important findings in our graph can be found in Figure~\ref{figure_subgraphs}.


\section{Conclusion}

This research aims to differentiate commercial and editorial content, and more specifically, advertorials from regular articles, and our main research questions are the following. To what extent can we differentiate advertorials and articles by using machine learning? And can we use machine learning and a data driven approach to better understand the difference between commercial and editorial language? 

We answered the first question by developing a range of models for various media, and deriving a lexicon from it. The best models perform with over 90 per cent accuracy, and as mentioned this is an optimistic estimate and performance clearly varies by medium and set up. Further insight is provided by highlighting the differences of performance across media, with business-to-business medium Ondernemer scoring lowest, which could make sense given similarities in jargon. Feature importance analysis and co-occurrence graphs provided further insight into differences in language, both from a topic perspective, as well as how these topics were being spoken about. 

Our research has some known limitations. In particular the size of the data set (of just 2000 entries) could be increased in future work, including a wider set of media and longer time frames. A key challenge here to overcome is that that particularly advertorials are not always available for extended periods of time. It may also be interesting to expand the scope to other major languages and other forms of native advertising. We also plan to engage with the general public, journalists as well as marketers, using the results of this research to raise awareness and trigger debate and discussion.

Despite some of it's limitations we think our research can serve as an example to put the problem on the agenda, provide insight into it, and illustrate the potential of using machine learning for differentiating commercial and editorial content. Moreover, it also showcases how machine learning and AI can be a solution, not just a problem, in society and the modern digital media landscape. 

%
%
%
\bibliographystyle{splncs04}
\bibliography{advertorials}
\end{document}